\documentclass[10pt,twocolumn,letterpaper]{article}

\usepackage{cvpr}              
\usepackage{marvosym} 

\definecolor{cvprblue}{rgb}{0.21,0.49,0.74}
\usepackage[pagebackref,breaklinks,colorlinks,allcolors=cvprblue]{hyperref}

\usepackage{graphicx} 
\usepackage{booktabs} 
\usepackage{makecell}   
\usepackage[utf8]{inputenc} 
\usepackage{rotating} 
\usepackage{tabularx} 
\usepackage{array}    
\usepackage{textcomp}
\usepackage{hyperref}

\def\paperID{44} 
\def\confName{CVPR}
\def\confYear{2025}

\title{Optimizing Multimodal LLMs for Egocentric Video Understanding: \\A Solution for the HD-EPIC VQA Challenge}       

\author{%
Sicheng Yang$^{\star}$, 
Yukai Huang$^{\star}$, 
Shitong Sun, 
Weitong Cai, 
Jiankang Deng, 
Jifei Song,
Zhensong Zhang\\[0.05ex]
\\[0ex]     
}

\begin{document}
\maketitle
\begin{abstract}
Multimodal Large Language Models (MLLMs) struggle with complex video QA benchmarks like HD-EPIC VQA due to ambiguous queries/options, poor long-range temporal reasoning, and non-standardized outputs. We propose a framework integrating query/choice pre-processing, domain-specific Qwen2.5-VL fine-tuning, a novel Temporal Chain-of-Thought (T-CoT) prompting for multi-step reasoning, and robust post-processing. This system achieves 41.6\% accuracy on HD-EPIC VQA, highlighting the need for holistic pipeline optimization in demanding video understanding. 
Our code, fine-tuned models are available at \href{https://github.com/YoungSeng/Egocentric-Co-Pilot}{here}.
\renewcommand{\thefootnote}{}

\end{abstract}

\section{Introduction}

Visual question answering (VQA) is a video understanding task that studies how to answer questions about video content based on its temporal visual information \cite{DBLP:journals/corr/abs-2405-21075, DBLP:journals/corr/abs-2502-04144}.
Compared to VQA for short, static or exocentric videos, implementing egocentric VQA for long videos \cite{DBLP:conf/cvpr/GraumanWBCFGH0L22, DBLP:journals/corr/abs-2502-04144, DBLP:journals/corr/abs-2406-18070, DBLP:journals/corr/abs-2501-19061} is significantly more challenging, as it requires robust temporal reasoning over long durations, inferring human intent, taking into account temporal information, history memory, and complex multistep reasoning for nuanced queries.        

To rigorously evaluate model performance on these specific, challenging egocentric VQA tasks, we focus on the VQA benchmark from the HD-EPIC dataset \cite{DBLP:journals/corr/abs-2502-04144}. 
HD-EPIC provides highly detailed long egocentric videos of complex kitchen activities, making its VQA benchmark a vital testbed for the aforementioned challenges. This benchmark is notable for its comprehensive coverage, featuring 30 distinct question prototypes that generate 26K questions specifically designed to test intricate temporal reasoning, object interaction, and fine-grained action understanding, which are crucial for evaluating models in this domain.

Despite the remarkable performance of many multimodal large language models (MLLMs) on general visual question answering benchmarks \cite{DBLP:conf/icml/ChiangZ0ALLZ0JG24}, including notable closed-source models like Gemini 2.5 Pro \cite{google2025gemini25pro}, GPT-4o \cite{DBLP:journals/corr/abs-2410-21276}, and Grok-3 \cite{xai2025grok3}, and prominent open-source counterparts such as DeepSeek-v3 \cite{DBLP:journals/corr/abs-2412-19437}, InternVL3 \cite{zhu2025internvl3}, and Qwen2.5-VL \cite{DBLP:journals/corr/abs-2502-13923}, they often exhibit limited performance on such specific, challenging tasks involving egocentric perspectives, complex kitchen scenarios, long temporal contexts, and intricate reasoning-based questions. 

\begin{figure}
    \centering
    \includegraphics[width=0.99\linewidth]{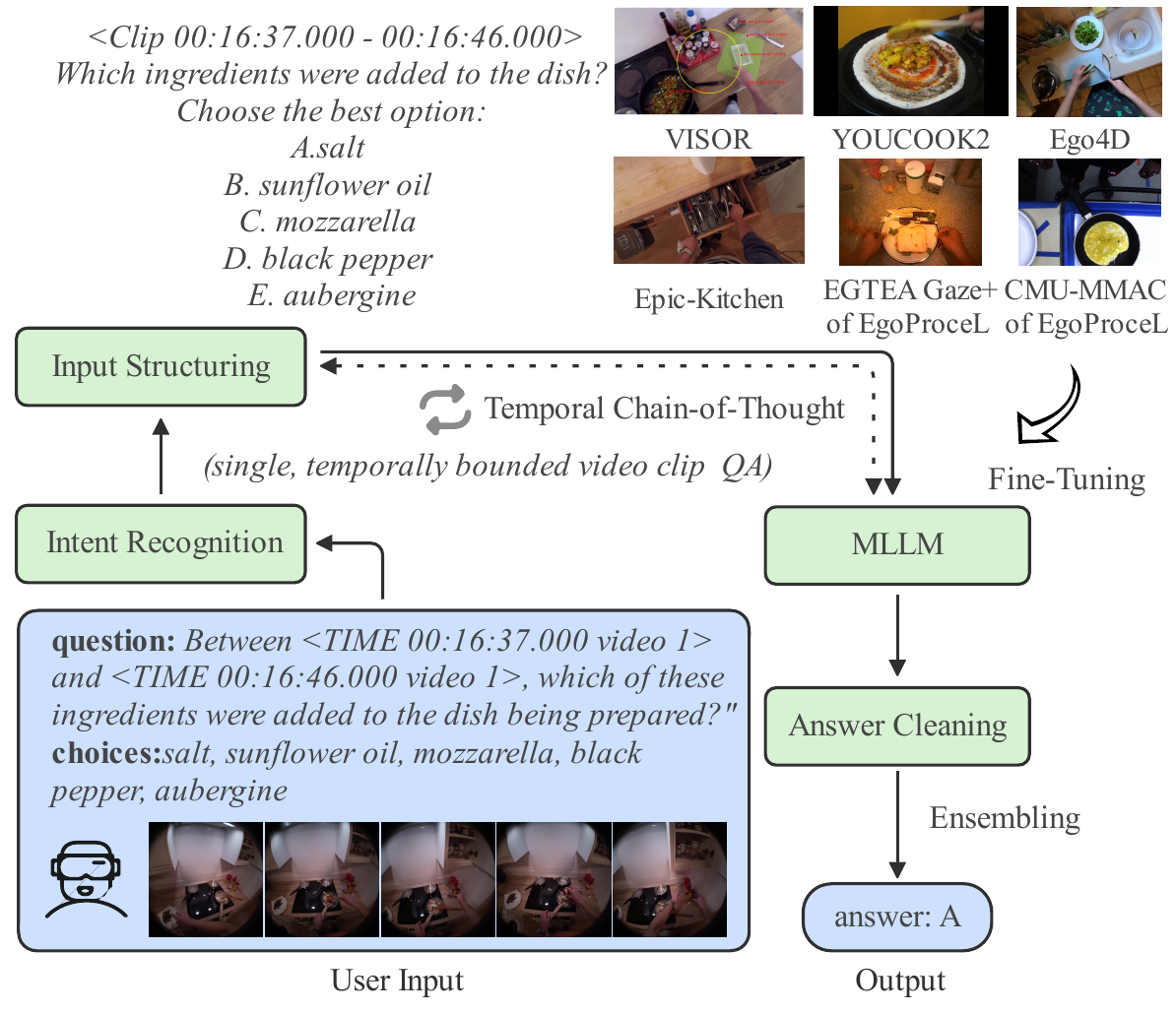}
    \caption{Overview of the proposed VQA system.}
    \label{framework}
\end{figure}

In this work, using Qwen-2.5 VL-7B \cite{DBLP:journals/corr/abs-2502-13923} as our base model, we analyze several factors critical to improving performance on this task as shown in Figure \ref{framework}: 
(1) Careful data pre-processing and optimization to better align user intent with MLLM comprehension. 
(2) Fine-tuning on extensive and densely annotated egocentric kitchen video data.
(3) Implementing a two-stage reasoning process for temporal questions, specifically for long videos and keyframes, enabling iterative processing and inference by the MLLM instead of a single-pass output.
(4) Robust post-processing, including handling anomalous outputs, result cleaning, and integrating responses from multiple prompts, which collectively led to our model exceeding the performance of previous baseline approaches.

\section{Method and Experiments}

\subsection{Question Intent Recognition and Clarification}       
\label{preprocess}

To enhance MLLM performance on specialized benchmarks like HD-EPIC VQA, meticulous pre-processing of queries and choices is paramount. We address the inherent ambiguity and varied formatting of the 30-question prototypes through a multi-faceted strategy for intent recognition and clarification.
First, we conduct a rigorous input modality analysis for each query, by classifying its visual context to the following 4 types:
(1) A single static image (often a keyframe, e.g., for \texttt{3d perception fixture location}).
(2) Multiple static images (e.g., for comparative tasks like \texttt{nutrition image nutrition estimation}).
(3) A single, temporally bounded video clip (e.g., \texttt{fine grained action recognition}).
(4) Multiple distinct video segments (e.g., \texttt{recipe prep localization}).
This classification critically informs subsequent processing tailored to the MLLM.


Secondly, we implement task-specific prompt refinement. Original questions are strategically transformed into clearer, structured formats that are more conducive to MLLM comprehension. Using regular expression-based parsing, we extract core entities, temporal markers, and relational constraints. These are then re-synthesized into improved prompts, for instance, by rephrasing concise queries (e.g., "where is X?") into more explicit, viewpoint-grounded questions (e.g., "Based on this image of my current viewpoint, determine the direction of X.").


Thirdly, based on the observation that MLLM is highly sensitive to the choice presentation  \cite{DBLP:journals/corr/abs-2502-04144}, we standardize and optimize the structure of multiple-choice options in the following ways. Converting numeric enumerators to alphabetic enumerators (e.g., A., B.) yielded an initial +1.6\% accuracy gain. Further subtle formatting—inter-option spacing or semicolons—improved accuracy by ~+1.8\% and ~+2.0\% respectively. Critically, clear newline delineation (\textbackslash n) for each option contributed a significant +2.4\% uplift. Complex temporal options with multiple time segments are also reformatted for clarity (e.g., A. [V1] HH:MM:SS.sss - HH:MM:SS.sss). These refinements, with uniform delimiters, reduce parsing ambiguity \cite{DBLP:journals/corr/abs-2511-08971}.

\begin{table}[!tbhp]
\resizebox{\columnwidth}{!}{%
\begin{tabular}{lcccccccc}
\hline
Model                 & Recipe & Ingredient & Nutrition & Action & 3D   & Motion & Gaze & Avg. \\ \hline
VideoLlama 2 \cite{DBLP:journals/corr/abs-2502-04144}         & 30.8   & 25.7       & 32.7      & 27.2   & 25.7 & 28.5   & 21.2 & 27.4 \\
LongVA  \cite{DBLP:journals/corr/abs-2502-04144}              & 29.6   & 30.8       & 33.7      & 30.7   & 32.9 & 22.7   & 24.5 & 29.3 \\
LLaVA-Video \cite{DBLP:journals/corr/abs-2502-04144}          & 36.3   & 33.5       & \textbf{38.7}      & \textbf{43.0}   & 27.3 & 18.9   & 29.3 & 32.4 \\
Gemini Pro \cite{DBLP:journals/corr/abs-2502-04144}           & 60.5   & \textbf{46.2}      & 34.7      & 39.6   & 32.5 & 20.8   & 28.7 & 37.6 \\
Qwen2.5 VL 7B In. \cite{DBLP:journals/corr/abs-2502-13923}    & 40.6       & 35.8           & 32.0          & 37.3       & 35.0     & 23.9       & 29.7     & 33.5     \\
Qwen2.5 VL 32B In. \cite{DBLP:journals/corr/abs-2502-13923}   & 59.0       & 37.0           & 33.0          & 40.3       & 35.6     & 19.8       & \textbf{33.6}     & 36.9     \\
Ours                  & \textbf{64.8}       & 43.3           & 37.0          & 42.0       & \textbf{40.9}     & \textbf{29.9}       & 33.0     & \textbf{41.6}     \\ \hline
w/o Pre-Processing      & 62.0       & 40.3           &  35.7         &  39.0      &  35.4    &  25.0      &  29.4    & 38.1     \\
w/o Fine-tuning / T-CoT & 62.6       & 41.5           & 32.0          & 38.3       & 35.2     & 23.5       & 29.1     & 37.5     \\
w/o Post-Processing     & 65.0       & 43.0           & 35.0          & 40.3       & 38.4     &  26.5      & 31.6     & 40.0     \\ \hline
\end{tabular}%
}
\caption{VQA Results per Category (\% Acc.). `w/o' is short for `without' in ablation study.}
\label{tab:my-table1}
\end{table}

These pre-processings enhance the MLLM's task comprehension by reducing cognitive load and aligning the input structure with the model's processing strengths. As shown in Table \ref{tab:my-table1}, these targeted refinements alone yield a discernible accuracy improvement (+3.5\%), highlighting robust input engineering as a vital precursor to advanced model capabilities.

\begin{table*}[htbp] 
\centering 

\newcommand{\rotAngle}{-45} 
\renewcommand{\theadfont}{\scriptsize}         

\newcommand{\rot}[1]{\thead[t]{%
    \raisebox{-18.5ex}{
        \turnbox{\rotAngle}{\makebox[0pt][r]{#1}}%
    }%
}}
\newcolumntype{Y}{>{\centering\arraybackslash}X}

\resizebox{\textwidth}{!}{
  \setlength{\tabcolsep}{6pt} 
  \scriptsize
  \begin{tabularx}{\linewidth}{@{}l*{30}{Y}@{}}
  \toprule 
  & 
\rot{Recipe Recognition} &
  \rot{Multi-Recipe Recognition} &
  \rot{Multi-Step Localization} &
  \rot{Step Localization
} & \rot{Prep Localization
} &
  \rot{Step Recognition
} & \rot{Rough Step Localization} &
  \rot{Following Activity Recognition} & \rot{Ingredient Retrieval} &
  \rot{Ingredient Weight} & \rot{Ingredients Order
} &
  \rot{Ingredient Adding Localization} & \rot{Ingredient Recognition} &
  \rot{Exact Ingredient Recognition} & \rot{Image Nutrition Estimation} &
  \rot{Nutrition Change} & \rot{Video Nutrition Estimation} &
  \rot{Action Recognition} & \rot{How Recognition} &
  \rot{Why Recognition} & \rot{Action Localization} &
  \rot{Fixture Location} & \rot{Object Location
} &
  \rot{Object Contents Retrieval
} & \rot{Fixture Interaction Counting
} &
  \rot{Object Movement Itinerary
} & \rot{Object Movement Counting
} &
  \rot{Stationary Object Localization
} & \rot{Gaze Estimation
} &
  \rot{Interaction Anticipation
} \\
  \midrule
  VideoLlama 2 \cite{DBLP:journals/corr/abs-2502-04144} & 22.0 & 52.0 & 18.0 & 38.0 & 13.0 & 13.0 & 21.0 & 64.0 & 19.0 & 30.0 & 20.0 & 27.0 & 26.0 & 32.0 & 24.0 & 20.0 & 54.0 & 30.9 & 25.2 & 32.2 & 20.7 & 18.8 & 31.0 & 35.5 & 17.7 & 11.0 & 44.0 & 30.5 & 30.0 & 12.4 \\
LongVA \cite{DBLP:journals/corr/abs-2502-04144}      & 14.0                 & 44.0                       & 36.0                        & 18.0 & 18.0 & 26.0 & 19.0 & 62.0 & 25.0 & 24.0 & 44.0 & 42.0 & 30.0 & 20.0 & 25.0 & 22.0 & 54.0 & 36.9 & 28.4 & 37.0 & 20.5 & 26.6 & 41.2 & 31.5 & 32.3 & 10.2 & 34.5 & 23.5 & 36.0 & 13.0 \\
LLaVA-Video \cite{DBLP:journals/corr/abs-2502-04144} & 28.0                 & 68.0                       & 44.0                        & 20.0 & 21.0 & 23.0 & 24.0 & 62.0 & 22.0 & 36.0 & 38.0 & 41.0 & \textbf{36.0} & 28.0 & \textbf{28.0} & \textbf{26.0} & \textbf{62.0} & \textbf{58.6} & \textbf{41.4} & \textbf{51.2} & 20.9 & 21.8 & 30.6 & 40.5 & 16.3 & 9.8  & 20.0   & 27.0   & 47.5 & 11.1 \\
Gemini Pro \cite{DBLP:journals/corr/abs-2502-04144}  & \textbf{42.0}                 & \textbf{76.0}                       & \textbf{88.0}                        & 70.0 & 35.0 & 45.0 & \textbf{74.0} & 54.0 & 49.0 & \textbf{46.0} & \textbf{56.0} & \textbf{62.0} & \textbf{36.0} & 28.0 & 26.0 & 16.0 & \textbf{62.0} & 49.3 & 35.6 & 43.2 & \textbf{30.3} & 20.8 & 32.4 & 41.5 & \textbf{35.3} & \textbf{18.0}   & 13.0   & \textbf{31.5} & 36.5 & \textbf{20.8} \\
Qwen2.5 VL 7B In. \cite{DBLP:journals/corr/abs-2502-13923}& 30.0                 & 64.0                       & 20.0                        & 20.0 & 20.0 & 78.0 & 21.0 & 72.0 & 71.0 & 28.0 & 34.0 & 22.0 & 30.0 & 30.0 & 20.0 & 20.0 & 56.0 & 50.1 & 33.8 & 48.4 & 16.9 & 27.0 & 40.0 & 46.5 & 26.3 & 11.2   & 38.0   & 22.5 & 45.4 & 14.0 \\
Qwen2.5 VL 32B In. \cite{DBLP:journals/corr/abs-2502-13923} & 28.0                 & 58.0                       & 62.0                        & \textbf{72.0} & 35.0 & 72.0 & 65.0 & \textbf{80.0} & 66.0 & 26.0 & 30.0 & 38.0 & 24.0 & \textbf{38.0} & 23.0 & 16.0 & 60.0 & 51.3 & 37.8 & 48.8 & 23.1 & 26.2 & 46.3 & 45.2 & 24.7 & 7.8   & 22.0  & 29.5 & 49.8 & 17.4 \\
Ours & 34.0                 & 72.0                       & 70.0                        & 68.0 & \textbf{50.0} & \textbf{81.0} & 73.0 & 70.0 & \textbf{76.0} & 32.0 & 36.0 & 48.0 & 30.0 & \textbf{38.0} & 25.0 & \textbf{26.0} & 60.0 & 55.6 & 36.6 & 51.0 & 24.9 & \textbf{34.2} & \textbf{49.8} & \textbf{50.5} & 29.0 &  14.2  &  \textbf{44.5} & 31.0 & \textbf{51.4} & 14.5 \\ \hline

w/o Pre-Processing & 30.0                 & 64.0                       & 66.0                         & 66.0 & 52.0 & 79.0 & 67.0 & 72.0 & 70.0 & 30.0 & 36.0 & 42.0 & 26.0 & 38.0 & 25.0 & 20.0 & 62.0 & 51.5 & 34.2 & 48.0 & 22.4 & 26.4 & 44.6 & 45.0 & 25.7 & 12.4   & 38.5   & 24.0 & 45.5 & 13.2 \\
w/o Fine-tuning / T-CoT & 26.0                 & 66.0                       & 66.0                        & 70.0 & 53.0 & 79.0 & 69.0 & 72.0 & 70.0 & 36.0 & 40.0 & 45.0 & 22.0 & 36.0 & 22.0 & 18.0 & 56.0 & 49.8 & 33.6 & 47.6 & 22.2 & 27.2 & 43.8 & 45.0 & 24.7 & 12.4   & 36.5   & 21.5 & 44.5 & 13.7 \\
w/o Post-Processing & 40.0                 & 68.0                       & 66.0                        & 70.0 & 52.0 & 81.0 & 69.0 & 74.0 & 74.0 & 30.0 & 40.0 & 44.0 & 30.0 & 40.0 & 27.0 & 22.0 & 56.0 & 52.2 & 35.4 & 48.6 & 24.8 & 30.0 & 47.6 & 46.5 & 29.3 &  14.0  & 40.0   & 25.5 & 47.4 & 15.8 \\
  \bottomrule
  \end{tabularx}%
}
\caption{Model results per question prototype. `w/o' is short for `without' in ablation study.}
\label{tab:my-table2}
\end{table*}
\subsection{Model Fine-Tuning and Temporal Domain Based Thinking}

Given that HD-EPIC \cite{DBLP:journals/corr/abs-2502-04144} is a first-person video dataset focused on kitchen scenarios, domain-specific fine-tuning is crucial for optimal performance \cite{DBLP:journals/corr/abs-2406-18070}. We fine-tuned the Qwen2.5-VL-7B-Instruct \cite{DBLP:journals/corr/abs-2502-13923} model on a diverse collection of egocentric kitchen video datasets, including EPIC-KITCHENS \cite{DBLP:journals/ijcv/DamenDFFKMMMPPW22, DBLP:journals/corr/abs-1804-02748, DBLP:journals/pami/DamenDFFFKMMPPW21}, CMU-MMAC \cite{de2009guide} and EGTEA Gaze+ \cite{DBLP:conf/eccv/LiLR18} of EgoProceL \cite{DBLP:conf/eccv/BansalAJ22} subsets, YOUCOOK2 \cite{DBLP:conf/aaai/ZhouXC18}, VISOR \cite{DBLP:conf/nips/DarkhalilSZMKHF22}, and selected portions of Ego4D \cite{DBLP:conf/cvpr/GraumanWBCFGH0L22} relevant to object interaction and procedural understanding. 
Fine-tuning was configured with a learning rate of $2 \times 10 ^{-7}$, batch size 2, 1 gradient accumulation step, and 1 epoch. Only LLM components were tuned, freezing the vision tower and MLP projector. We utilized bfloat16 precision, the AdamW optimizer, and a maximum sequence length of 131072 tokens. Video processing involved a maximum of 768 frames/sample (minimum 4), with dynamic total pixel adjustment per video (3136 to 846720).

Despite fine-tuning, Qwen2.5-VL, like many MLLMs, exhibited challenges with long-term temporal relationship understanding. For instance, on \texttt{Multi-Step} \texttt{Localization}, our fine-tuned model achieved 26\% accuracy (vs. 22\% pre-tuning), substantially below Gemini Pro's 88\%. Similar disparities occurred in \texttt{Step} \texttt{Localization} (25\% vs. 70\%) and \texttt{Rough} \texttt{Step} \texttt{Localization} (28\% vs. 74\%), and tasks like \texttt{Ingredients} \texttt{Order}. We hypothesize this stems partly from the model's training sequence lengths (8192/32768) and its video processing strategy, capping analyzed frames at 768 (total video tokens $\leq$ 24576). For videos $>$12 minutes (at 1 FPS, $>$720 frames), dynamic resolution scaling or token limits may yield insufficient effective input frames for fine-grained temporal analysis.

To address these temporal limitations, we developed a Temporal Chain-of-Thought (T-CoT) prompting strategy \cite{DBLP:conf/nips/Wei0SBIXCLZ22, wang2025chainofmodality}, guiding the MLLM through intermediate reasoning steps to isolate and comprehend relevant temporal context, rather than directly posing complex temporal queries. This strategy encompasses:
(1) Explicit Temporal Cue Exploitation: For tasks with localized visual information or specified time points/segments (e.g., \texttt{3d perception} with bounding box (BBOX), \texttt{gaze} with time segments), we first process these cues. BBOX information is resolved by prompting the MLLM to generate a noun phrase for the object within the BBOX, which then replaces the BBOX placeholder in the question. For specified timestamps or segments, we extract the relevant clip and prompt the MLLM to analyze or narrate its content.
(2) Focused Temporal Windowing: For questions implicitly tied to a narrow temporal window around a key event (e.g., \texttt{3d perception object location} implying ``now''), we dynamically segment the video to a shorter duration (e.g., $\pm$ 10s around the relevant point), focusing MLLM attention and reducing irrelevant processing.
(3) Multi-Video Synchronization: When questions or options involve multiple distinct video clips (e.g., \texttt{recipe prep localization}), these are programmatically concatenated. All timestamps in the question and options are then re-normalized relative to this new unified timeline, enabling the MLLM to process a single, coherent video stream.
(4) Hierarchical Processing for Long Videos: For tasks requiring detailed understanding of extended videos exceeding the MLLM's single-pass capacity (e.g., \texttt{ingredient ingredients order}), we employ a chunking strategy. The video is divided into manageable, non-overlapping segments (e.g., 10-min, $<$ 768 frames). The MLLM generates a concise narration for each chunk. These temporally ordered narrations are then aggregated and prepended to the original question, providing rich, summarized contextual background for the final MLLM reasoning.

Our proposed two-stage T-CoT process—initially extracting, segmenting, or summarizing temporal/spatial context, followed by addressing the VQA query with this refined input—substantially reduces MLLM cognitive load and irrelevant information. Our T-CoT approach yielded a +3.0\% overall accuracy improvement across all tasks compared to direct VQA with only initial pre-processing, demonstrating its efficacy in enhancing MLLM reasoning for complex temporal video understanding.
We report results for the 7 categories HD-EPIC VQA scores (Table \ref{tab:my-table1}) and 30 task details (Table \ref{tab:my-table2}), and found that the fine-tuning and T-CoT strategies had the greatest impact on the results.


\subsection{Answer Cleaning and Ensembling}       


MLLMs, despite explicit instructions for single-letter outputs (e.g., A-E), may generate verbose responses, hindering automated evaluation. We introduced a robust post-processing step via an answer cleaning \cite{DBLP:journals/tois/DiSWML25} module. This module employs regular expressions to parse raw MLLM textual outputs, extract the most probable single-letter choice, and convert it to a zero-based index for ground truth comparison. This cleaning procedure proved crucial, enabling automated scoring, mitigating misinterpretations of verbose outputs, and improving accuracy by +0.8\% over evaluating raw outputs.


To enhance prediction robustness and accuracy for the multiple-choice HD-EPIC VQA benchmark, we implemented an ensembling strategy. This involved generating five distinct, semantically equivalent prompts per question by subtly varying the phrasing while preserving core semantic elements (entities, temporal information, relational constraints from Section \ref{preprocess}). The MLLM processed each prompt independently, and the final answer was determined via majority voting over the five cleaned predictions.






\section{Discussion and Conclusion}


Our comprehensive strategy—unifying input pre-processing, domain-specific fine-tuning, Temporal Chain-of-Thought (T-CoT) prompting, answer cleaning, and ensembling—demonstrably elevates MLLM performance on the HD-EPIC VQA benchmark.
Employing Qwen2.5-VL-7B-Instruct, we observed that direct scaling to its 32B variant offered no proportional performance gains within our current pipeline. This is attributed to larger models' increased verbosity or uncertainty when constrained, and their extended T-CoT generations introducing noise detrimental to reasoning (e.g., an overabundance of detailed short video segments proved less effective than fewer, longer ones). Targeted fine-tuning of these larger 32B or 72B models is posited as a more promising path.
Despite significant baseline improvements, our multi-stage architecture incurs latency, posing a critical performance-efficiency trade-off, particularly for real-time applications. Optimizing this balance is imperative for future work.
Furthermore, a substantial gap persists towards human-level cognition \cite{taluzzi2025pixels}, notably in tasks requiring deep reasoning and robust long-term memory for extended videos, underscoring a crucial research trajectory.

In conclusion, we presented a comprehensive methodology that demonstrably enhances MLLM efficacy for complex egocentric video understanding. Our results underscore the collective importance of structured input/output processing, domain adaptation, and guided temporal reasoning. Despite persistent challenges in computational efficiency and attaining human-level cognition, our work provides a robust baseline and crucial insights for the future development of advanced and practical AI systems targeting egocentric video analysis.


{
    \small
    \bibliographystyle{ieeenat_fullname}
    \bibliography{main}
}


\end{document}